\documentclass[preprint,12pt]{elsarticle}

\usepackage{amssymb}
\usepackage{caption}
\usepackage{multirow}
\usepackage{booktabs}
\usepackage{amsmath}
\usepackage{makecell}
\usepackage{adjustbox}
\usepackage{threeparttable}
\usepackage{algorithmic}
\usepackage{algorithm}
\usepackage{url}

\biboptions{numbers,sort&compress} 

\begin{document}

\begin{frontmatter}


\title{NAS-ASDet: An Adaptive Design Method for Surface Defect Detection Network using Neural Architecture Search}

\author[1]{Zhenrong Wang}
\author[1]{Bin Li\corref{mycorrespondingauthor}}
\ead{libin999@hust.edu.cn}
\author[1]{Weifeng Li}
\author[1]{Shuanlong Niu}
\author[1]{Wang Miao}
\author[1]{Tongzhi Niu}

\cortext[mycorrespondingauthor]{Corresponding authors}
\address[1]{ School of Mechanical Science and Engineering, Huazhong University of Science and Technology, Wuhan 430074, China}

\begin{abstract}
	Deep convolutional neural networks (CNNs) have been widely used in surface defect detection. However, no CNN architecture is suitable for all detection tasks and designing effective task-specific requires considerable effort. The neural architecture search (NAS) technology makes it possible to automatically generate adaptive data-driven networks. Here, we propose a new method called NAS-ASDet to adaptively design network for surface defect detection. First, a refined and industry-appropriate search space that can adaptively adjust the feature distribution is designed, which consists of repeatedly stacked basic novel cells with searchable attention operations. Then, a progressive search strategy with a deep supervision mechanism is used to explore the search space faster and better. This method can design high-performance and lightweight defect detection networks with data scarcity in industrial scenarios. The experimental results on four datasets demonstrate that the proposed method achieves superior performance and a relatively lighter model size compared to other competitive methods, including both manual and NAS-based approaches.
\end{abstract}
\end{frontmatter}


\section{Introduction}\label{sec1}

In industrial production, defects inevitably occur on the end-product surface. Surface defect detection is an effective method to control the quality of industrial products. With the rapid development of intelligent manufacturing, automated and intelligent defect detection has gradually become critical~\cite{wen2022steel}.

Recently, deep convolutional neural networks (CNNs) have been proven effective in surface defect detection. By designing different neural networks, improved detection performance is obtained.
For example, \cite{ni2021attention} and \cite{zhuxi2022lightweight} enhance the ability to focus on critical defects in complex semantics by incorporating attention modules. \cite{chen2023dcam} and \cite{yao2022ayolov3} design deformable convolution and dilated convolution, respectively, to improve the detection of irregular and diverse defects by expanding the receptive fields. \cite{luo2023ultrasmall,mrd2021,yang2019multiscale} use multi-scale feature fusion to integrate information from different levels, enhancing the awareness of defect scale variations. However, industrial scenarios encompass a wide range of surface defect types, each with significantly different characteristics~\cite{pga2019}. There is no unified design paradigm for how to design effective network matching data characteristics.

The design of traditional detection networks often relies on human expertise and repetitive experimentation, with certain difficulties and challenges~\cite{elsken2022neural}. Firstly, the design process lacks automation and optimization capabilities. Manual network design and adjustments require extensive trial and error, consuming a lot of manpower, time, and computational resources. Secondly, due to limitations in human cognition, traditional designs usually rely on predetermined network connectivity (VGGNet \cite{simonyan2014very}, ResNet~\cite{he2016deep},~etc), which may limit the ability to exploit feature information, resulting in suboptimal performance. Moreover, surface defects are diverse and the features of different types of defects are quite different. It is difficult for a single network to show general good performance on different complex detection tasks, and it needs to rely on additional manual design and adjustment.

To overcome these shortcomings, we introduce neural architecture search (NAS) into the network design for surface defect detection. NAS can automatically design data-driven networks to adapt to diverse requirements, so it can improve the efficiency of network design and make the searched network have excellent performance.
At present, NAS has made many breakthroughs in the field of natural imaging. Early NAS research search the entire network from scratch. The resultingly huge resource consumption limited the development of NAS. Therefore, researchers improved NAS with respect to search space and search strategy. Regarding the search space, NASNet \cite{nasnet2018} and NAS-Unet \cite{nasunet2019unet} use repeated cells to limit the search space size, which reduced the search difficulty. LiDNAS \cite{huynh2022lightweight} built the search space on predefined backbone networks to balance layer diversity and search space size. Regarding the search strategy, by adding the weight sharing mechanism, the gradient descent search strategy reduces the search cost.

Although NAS has been successfully applied in natural scenarios, there are few reports in the literature on the performance of NAS for surface defect detection network design in complex industrial scenarios. When applying NAS technology to surface defect detection network design, it is important to focus on unique industrial requirements, distinct from natural scenarios.

\begin{enumerate}
	\item{Limited available defect samples: The nature of industrial production lines leads to the generation of limited defective samples, which poses challenges to the search process. To overcome this issue, one solution is to use a reduced search space to decrease data requirements \cite{verma2023rnas}. However, due to restricted layer diversity, using reduced search spaces designed for natural scenarios may restrict the expressive power and potentially overlook excellent architectures for defect detection tasks. Therefore, NAS methods for surface defect detection should use a small search space and give careful consideration to defect features, focusing on structures and parameters which relevant to the defect characteristics to strike a balance between search space size and expressive capability.}
	
	\item{Stable detection accuracy and robustness: Unlike natural image segmentation, which focuses more on overall scene understanding, surface defect detection requires stable precise localization and identification of boundary contours to ensure product quality. Additionally, the production line introduces environmental disturbances like lighting variations, noise, and occlusion, making it necessary for the detection capability to be robust. Therefore, NAS methods need to fully consider the challenges in surface defect in order to adaptively design the network that meets the requirements.}
	
	\item{Lightweight and low-computation: Detection equipment on industrial production lines typically faces constraints, including limited memory space and constrained computational resources. Therefore, designing a low-computation and lightweight network is a crucial aspect of industrial network. To meet the requirements of detection accuracy within these resource limitations, the surface defect detection network designed by NAS should possesses lightweight characteristics while ensuring network performance.}
\end{enumerate}

To achieve the aforementioned requirements, we propose a NAS method specifically tailored for designing surface defect detection networks. Our approach considers both the search space and search strategy. Regarding the search space, we design a refined and industry-appropriate search space that enables NAS has good network design ability with limited defect samples. This search space enables data-driven design of lightweight detection networks that robustly adapt to various surface defect scenarios while balancing accuracy and computation. Additionally, we incorporate prior knowledge from manually designed detection networks into the propose NAS framework, enhancing the accuracy and robustness of the searched networks for different detection tasks. This involves the use of large receptive field cells and searchable attention operations to improve adaptability in complex environmental conditions, as well as a multi-scale feature fusion structure that can adaptively adjust the feature distribution to handle diverse shapes and scales. As for the search strategy, we enhance the efficiency of the gradient optimization search strategy (DARTS), so that the search space can be explored more efficiently, ensuring a performance-driven and time-efficient network design process.

The contributions of this article are as follows:
\begin{itemize}
	\item We propose a method to adaptively design the surface defect detection networks based on NAS, called NAS-ASDet. This scheme has a refined and industry-appropriate search space, which can adaptively search the lightweight defect detection network in industrial scenarios with limited data (compared to natural scenarios), reducing the workload of manual detection network design.
	
	\item A basic cell containing multiple receptive fields with searchable attention operations are provided to construct the search space, improve the detection ability of irregular and diverse defects, and enhance the ability to automatically focus on key defects in complex environments.
	
	\item A multi-scale feature fusion that can adaptively adjust the feature distribution is designed in the proposed NAS framework, enhancing the network's adaptability to defects of various scales and further improving the detection accuracy.
	
	\item We design a progressive search strategy with a deep supervision mechanism based on gradient optimization search strategy to effectively explore the search space. This strategy can make the search process better and faster to adjust the architecture to adapt to the defects to be detected, and further improve the efficiency of network design.
	
	\item The proposed method is capable of searching for networks with state-of-the-art resutls on four different surface defect datasets. Compared to recent manually detection networks, NAS-ASDet utilizes only approximately 10\% of the parameters and 5\% of the FLOPs but achieving the best performance. This proves that proposed NAS method has certain generality and theoretical value.
\end{itemize}

The rest of this article is organized as follows. Section~\ref{sec2} briefly introduces the related research work, including CNN-based methods for surface defect detection and the development of NAS methods. The proposed method for adaptive surface defect detection is described in Section~\ref{sec3}. Section~\ref{sec4} presents the experiments and discussions. Conclusions are given in Section~\ref{sec5}.

\section{Related work}\label{sec2}
\subsection{CNN-based Surface Defect Detection}
Since pixel-level defect detection can describe the defect contour boundary, which provides a more valuable reference for the evaluation of defect severity, segmentation maps are often used for defect localization. Inspired by FCN \cite{fully2015}, many methods (U-Net \cite{ronneberger2015u}, PSPNet \cite{zhao2017pyramid}, DeepLab\cite{chen2018encoder},  etc.) try to extract more informative features from image patches, leading to successful segmentation results. However, the performance of algorithms is often limited by the complex and diverse defect types in industrial scenes. Therefore, in recent years, defect segmentation methods have been adapted according to specific application scenarios to achieve improved detection performance.

Addressing the challenges of fuzzy boundaries (low contrast) and noise interference, 
DCAM-Net \cite{chen2023dcam} used deformable convolution and attention mechanism to locate the contrast of irregular strip surface defects. TSERNet \cite{han2022two} employed a prediction and refinement strategy, using edge information twice to generate saliency maps with more accurate boundaries and precise defect positions for steel strip defects. 
Addressing the challenges of multi-scale variations of inspected defects, MRD-Net \cite{mrd2021} captured short and long distance patterns through a multiscale feature enhancement fusion and reverse attention network. CSEPNet \cite{ding2022cross} designed a cross-scale edge purification network to highlight defects in steel images and maintain important edge information. 
Addressing the challenge of detecting small defect targets, \cite{luo2023ultrasmall} realized the ultrasmall bolt defect detection through feature fusion, attention mechanisms, and extraction of fused salient regions. \cite{li2023deep} improved the detection effect of small targets in wire and arc additive manufacturing by attaching a multi-SPP structure to the FPN. 
In addition, some studies such as FHENet \cite{zhou2023fhenet} and BV-YOLOv5S \cite{du2022improvement} focus on the design of lightweight detection networks to address the deployment challenges of large models in industrial scenarios.

Although these methods achieve good performance, they still require improvement: 1) They are usually based on a predetermined network connectivity. This means that features are extracted restrictively and fixedly, rather than being driven by data features, which limits performance. 2) Even though some methods try to design from scratch, they are usually customized for specific tasks. There is no single network that has shown competency for all detection tasks. Therefore, designing effective networks for specific tasks consumes considerable time and computation resources.

\subsection{Neural Architecture Search (NAS)}
NAS aims to design the neural architecture in an automatic way to maximize performance while using limited computing resources.

\subsubsection{Gradient Optimization-Based NAS} \label{subsub2-1}
The differentiable search method converts a discrete search space into a continuous differentiable form such that gradient descent can be applied to the search process to exceed the black-box optimization efficiency, such as reinforcement learning and evolutionary algorithms. The earliest gradient-based idea was proposed in DARTS \cite{darts2018}. 
But there remain two problems \cite{ren2021comprehensive} in gradient-based DARTS: (a) Full candidate operations participate in the whole search process, which leads to longer search times and heavier computational overhead; (b) The transfer of rough decoupling may leads to performance variation.
Many studies have improved DARTS: DATA \cite{data2019} developed the ensemble Gumbel softmax estimator, which realized migration between the search and validation stages. PC-DARTS \cite{pcdarts2019} proposed channel sampling and edge normalization technologies to reduce GPU resource consumption. P-DARTS \cite{pdarts2019} considered performance collapse and adopted a progressive search strategy. Even so, stably and efficiently searching strategies remain a popular research topic.

\subsubsection{Neural Architecture Search for Image Segmentation}\label{subsub2-2}
Auto-DeepLab \cite{liu2019auto} is generally recognized as the pioneering work demonstrating NAS application in image segmentation. It also extended the gradient optimization-based strategy to the segmentation task for the first time.
NAS-Unet \cite{nasunet2019unet} improved on the baseline U-net using NAS and showed higher performance.
FasterSeg \cite{fasterseg2019} used a multibranch architecture to overcome model breakdown.
DCNAS \cite{dcnas2021} designed a more complex supernet than Auto-DeepLab, which added cross-layer connections to the search space.
DNAS \cite{dnas2022} designed a three-level decoupled search strategy to reduce the training difficulty.
In addition, in order to eliminate the deployment pressure of large networks, some methods \cite{gu2021inas,yan2021lighttrack,vu2023toward,yao2023lightweight} also take into account the combined effect of NAS and hardware awareness to search for lighter architectures. However, most of the current NAS methods are designed based on natural scenes.

Although there have been some recent attempts of NAS in industrial scenarios, most of them are limited to defect classification or focused on specific tasks. For example, \cite{chen2021depth} and \cite{chen2022study} realized defect classification of steel cracks using NAS. \cite{shi2020defect} explored an automated defect detection method for industry wood veneer, and \cite{zhang2023lightweight} developed a NAS-based detection approach for analyzing defects in photovoltaic cells in electroluminescence images. Such methods do not meet the requirements of robustly designing precise defect localization in diverse industrial scenarios.

Therefore, the design method of NAS pixel-level defect detection network with high performance in industrial scenarios remains largely unexplored.

\section{Proposed method}\label{sec3}
\subsection{System Overview}
\begin{figure}[h]%
	\centering
	\includegraphics[width=10cm]{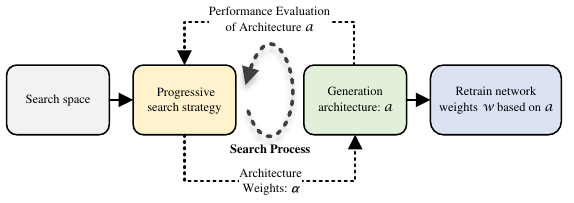}
	\caption{The framework for network architecture design in NAS-ASDet.}\label{Fig1}

\end{figure}

Considering the difficulty of manually designing detection networks and the challenges posed by the unique industrial requirements to NAS, we focus on the surface defect detection and propose an adaptive network design method, called NAS-ASDet.
In this method, surface defect detection is regarded as a pixel-level segmentation task.
NAS-ASDet allows the network to adaptively adjust the connection mode according to the data characteristics, and finally obtain a certain detection network with both high performance and lightweight. The adaptive network design process of the proposed method is shown in Figure.~\ref{Fig1}.

\begin{itemize}
	\item First, the refined and industry-appropriate lightweight search space is defined. 
	It is composed of repeatedly stacked expressive basic cells, where the cells consist of a given set of candidate operations (e.g., convolution, pooling). Correspondingly, the search scope is limited to the structure of the cell rather than the whole network.
	
	\item Next, the lightweight network architecture is searched. A progressive search strategy is used to explore the search space gradually. To make gradient optimization applicable, each candidate operation is assigned an updatable architectural weight $\boldsymbol{\alpha}$, and the cell consists of candidate operations with weight assignments. The cell performance is fed back to update $\boldsymbol{\alpha}$. According to the contribution of the candidate operation to performance, the weak operations are progressively removed so that the basic cells with favorable performance are obtained. 
	
	\item Then, the original supernet is replaced by the searched cells to obtain a definite architecture. On this basis, the network weights $w$ of the determined architecture are retrained to ensure complete convergence. 
	
	\item Finally, the fully trained deterministic lightweight network is used to achieve end-to-end defect segmentation and then to evaluate the detection performance. 
\end{itemize}

The main contributions of the above NAS process are two core contents: (a) The design of the refined and industry-appropriate lightweight search space, including the basic cell containing multiple receptive fields with searchable attention operations and the network architecture with adaptively fused multiscale features; (b) The design of a progressive search strategy with a deep supervision mechanism, where the search process is divided into multiple stages to explore the search space better and faster.

\begin{figure*}[!b]
	\centering
	\includegraphics[width=1.05\textwidth,height=0.4\textheight]{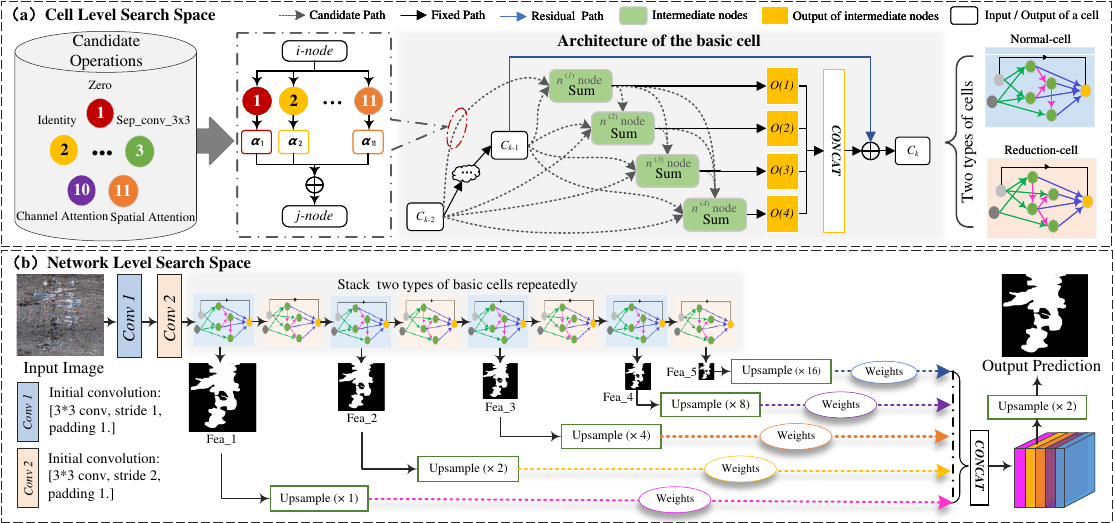}
	\caption{Schematic view of the proposed detection architecture search space in NAS-ASDet, which contains cell-level and network-level search spaces. The normal cell and reduction cell are the basic building modules of the network architecture. (a) In the cell-level search space, the basic cell consists of the candidate operation set, and the blue line represents the residual connection. The cell connection mode is automatically determined by the search process. (b) In the network-level search space, the detection architecture consists of repeatedly stacked basic cells, and the multiscale features are adaptively adjusted to produce the final prediction.
	}\label{Fig2}
\end{figure*}

\subsection{Search Space Design}
This section describes the refined and industry-appropriate lightweight search space designed for surface defect detection, which defines the architecture set that can be represented in theory. 
Considering the data scarcity in industrial scenarios, a small search space size should be used, inspired by~\cite{psp2022}, we use a reduced search space based on repeatedly stacked cell.
Specifically, this cell-based search space includes two levels: cell-level and network-level, as shown in Figure.~\ref{Fig2}. At the cell level, we design the lightweight cell containing multiple receptive fields with searchable attention operations to enrich the candidate operations. At the network level, we propose a new network-level framework with a multi-scale feature fusion that can adaptively adjust the feature distribution.

\subsubsection{Cell Level Search Space}
Since a small search space size should be used to reduce search difficulty, noting that many handcrafted architectures are based on repeated fixed blocks and inspired by \cite{nasnet2018}, we desgin the search space for the lightweight network based on two types of repeatedly stacked lightweight cells: normal cells and reduction cells. Both of these cells follow the same construction pattern, wherein the output feature map of normal cell contains the same dimension as the input and half the size of reduction~cell.

Many manually designed detection networks enhance their ability to detect defects in complex, irregular, and diverse contexts by using stronger multiscale representation. Therefore, inspired by Res2Net \cite{gao2019res2net} to expand the receptive field, we design the cell to generate multiple available receptive fields at a finer cell level granularity. The cell-level search space is shown in Figure.~\ref{Fig2}(a).

As the industrial scene is more seriously affected by complex environments and lighting changes, the  manual detection network usually adds attention mechanism to enhance the ability to extract important feature information, but when and how to add attention operation still requires researchers to analyze specific problems. Therefore, we propose to use attention operations to enrich the candidate operations, including channel attention \cite{hu2018squeeze} and spatial attention \cite{cbam2018}. So that cells can adaptively focus on key defects in complex environments.

Since industrial scenarios have lightweight and low-computation requirements, we first collect basic operations that are widely used in detection networks (such as convolution, pooling, etc.), and then use separable convolutions instead of ordinary convolutions to limit cell lightweighting. And the final full candidate operation set $\mathcal{O}$ is summarized in Table~\ref{Tab1}.

\begin{table}[h]
	\vspace{-0.1cm}
	\setlength{\baselineskip}{0.5em}
	\caption{\centering Candidate operation set in NAS-ASDet\label{tab 1}} 
	\centering 
	\resizebox{85mm}{27mm}{
		\begin{tabular}{c c c} 
			\toprule 
			ID & Operation & Note \\
			\midrule	
			1 &Zero & no operation\\
			2 &Identity & skip connection\\
			3 &Sep\_conv\_3x3 & 3x3 separable convolution\\
			4 &Sep\_conv\_5x5 & 5x5 separable convolution\\
			5 &Sep\_conv\_7x7 & 7x7 separable convolution\\
			6 &Dil\_conv\_3x3 & 3x3 dilated separable convolution \\
			7 &Dil\_conv\_5x5 & 5x5 dilated separable convolution\\
			8 &Max\_pool\_3x3 & 3x3 max pooling\\
			9 &Avg\_pool\_3x3 & 3x3 average pooling\\
			10 &Channel\_att & channel attention\\
			11 &Spatial\_att & spatial attention\\
			\bottomrule 
		\end{tabular}\label{Tab1}}
\end{table}

Each cell is viewed as a directed acyclic graph consisting of multiple edges and $N$ nodes. Two of nodes ($C_{k-1}$,$C_{k-2}$) are inputs, one ($C_{k}$) is output, and the remaining $N-3$ are intermediate nodes. The transformation from the $i$-node to the $j$-node is connected by a mixed edge, which is a mixture of all candidate operations denoted $O^{(i,j)}$. To enable gradient optimization, the selection of discrete operations is relaxed as a continuous optimization problem. Specifically, we assign each candidate operation in each edge with an architecture weight $\alpha_o(\alpha_o \in \boldsymbol{\alpha})$, and $softmax$ is applied to ensure that $\alpha_o\in [0,1]$, $\sum \alpha_o = 1 $. The information flow between the $i$-node and $j$-node can be calculated:
\begin{equation}
	\overline{o}^{(i,j)}(x_i)=\sum_{o \in \mathcal{O}}\frac{e^{\alpha_o^{(i,j)}}}{\sum_{o' \in \mathcal{O}}e^{\alpha_{o'}^{(i,j)}}}o(x_i),~~(i<j)
\end{equation}
where $i<j$ represents that $i$ is the forward node of $j$, and the data flow needs to be transmitted from $i$ to $j$. $x_i$ represents the $i$-th node feature map in cells, and each candidate operation $o(\cdot)$ belongs to candidate operation set space $\mathcal{O}$ with a relative weight $\alpha_o^{(i,j)}$. $\overline{o}^{(i,j)}(x_i)$ represents the feature map corresponding to node $j$ after data is transmitted from node $i$ to node $j$.

Each intermediate node accepts all forward intermediate node feature maps and the output of the previous two cells:
\begin{equation}
	n^{(j)}=\sum_{i<j}\overline{o}^{(i,j)}(n^{(i)})+\sum_{m=k-2}^{k-1} \overline{o}^{(m,j)}(C_m)
\end{equation}
where $n^{(i)}$ represents the feature map of the $i$-th intermediate node, $C_m$ denotes the forward cell output, and $m=k-1,k-2$ indicate the previous and previous-previous cells, respectively.

Then, the cell concatenates all intermediate nodes and the residual feature map sum as its output:
\begin{equation}
	C_k = concat~(n^{(i)})+res(C_{k-1}),~~i \in \{1,\cdots,N-3\}
\end{equation}
where $C_k$ represents the output of the $k$-th cell.

Finally, the architecture $\mathcal{A}$ with candidate operation set $\mathcal{O}$ can be formed by stacking multiple cells:
\begin{equation}
	\mathcal{A}(\mathcal{O}) \cong \{\overrightarrow{C_{nor},C_{red},\cdots,C_{nor},C_{red}} \}
\end{equation}
where $C_{nor}$ and $C_{red}$ denote the normal cell and reduction cell. $\{\overrightarrow{\cdot} \}$ represents the direction of information transmission.

This lightweight cell level search space provides several guarantees for detection performance: 1) Instead of using fixed connection and operations, the cell can be automatically searched and adaptively select an appropriate connection mode, such that more appropriate fine-grained features can be obtained and multiple available receptive fields can be provided, which improves the ability to detect defects. 2) Due to the addition of forward node $C_{k-2}$, the network can choose to accept more forward information and expand the receptive field. 3) Attention operations enrich the candidate operation set, and it is automatically determined by search, which enhances the ability to automatically focus on key defects in complex environments.

\subsubsection{Network Level Search Space}
The segmentation NAS pioneered by Auto-DeepLab \cite{liu2019auto} usually uses single-level~features for target prediction. However, the defect scales are~varied, and single-layer features offer limited ability to recognize different scale targets. Noting that the effectiveness of combining multiscale features for performance improvement has been demonstrated in handcrafted architectures, 
we are motivated to integrate these effective knowledge of multiscale feature fusion into the proposed NAS framework to enhance the ability of the network to deal with multi-scale challenges, as shown in Figure.~\ref{Fig2}(b).

It should be noted that although handcrafted network performance proves that multiscale feature combination is effective, it does not indicate that in FPN-based detection architectures, balanced multiscale feature fusion is best. Different surface defects present different shapes and scales, which signifies that different level features have different levels of importance for detection, e.g., shallow features are more important for small-scale detection, and vice versa \cite{mrd2021}.
Therefore, we assign different weights at different feature levels, which are learned during the training process, toward adaptively adjusting the importance distribution of features and focusing on more useful information. The implementation details are as follows.

First, for given defect images, multilevel features are extracted by repeatedly stacking cells. The last layer features of each level $\{Fea_1,Fea_2,\cdots,Fea_5\}$ are used for subsequent adaptive feature fusion. To integrate feature maps on different scales, these maps are resized to $Fea_1$:
\begin{equation}
	\overline{Fea_i}=upsample^n(Fea_i),~~ i \in \{1,2,3,4,5\}
\end{equation}
where $upsample^n(\cdot)$ indicates the upsampling operation by enlarging features by $n$ times, and $n \in \{1,2,4,8,16\}$. We concatenate $\overline{Fea_i}$ in the channel dimension to obtain the original feature fusion map $\mathcal{F}$:
\begin{equation}
	\mathcal{F}=concat(\overline{Fea_1},\overline{Fea_2},\overline{Fea_3},\overline{Fea_4},\overline{Fea_5})
\end{equation}

Next, learnable weights are added to $\mathcal{F}$, which are used to adjust the importance distribution of different features. Specifically, we use the average pooling operation to obtain the representation vectors $V$ of each layer in $\mathcal{F}$. The fully connected layer $FC$ and nonlinear activation function $ReLU$ are used to encode $V$ to obtain $V'$:
\begin{equation}
	\begin{aligned}
		V&=avg\_pooling(\mathcal{F}), \\
		V'&=ReLU(FC(V,w_{\mathcal{F}})),
	\end{aligned}
\end{equation}
where $w_{\mathcal{F}}$ represents the parameter of $\mathcal{F}$.

Then, $V'$ is normalized to $[0, 1]$ through the $sigmoid$ function and used to represent feature importance:
\begin{equation}
	V_f = sigmoid(V')
\end{equation}

Therefore, network can adaptively adjust the feature importance by learning $V_f$. The enhanced features can be expressed:
\begin{equation}
	\mathcal{F}_{enhance}=V_f \times \mathcal{F}
\end{equation}

Finally, we resize the feature map in $\mathcal{F}_{enhance}$ using an upsampling operation to complete end-to-end segmentation prediction:
\begin{equation}
	Output =upsample^2(\mathcal{F}_{enhance})
\end{equation}
Thus, the lightweight network-level search space design is completed.

The network level search space considered multiscale features inspired by handcrafted architectures, which increases the lightweight cell-based search space expression. The generated architecture can not only extract data-driven multiscale features but also adaptively adjusts the feature's importance distribution to cope with scale challenges, which guarantees effective detection under broadly diverse and complex factors.

\subsection{Search Strategy}\label{sub3-3}
\begin{figure}[!h]%
	\centering
	\includegraphics[width=10.5cm]{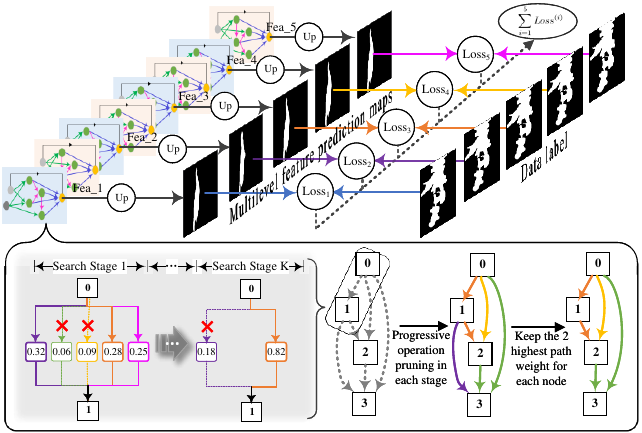}
	\caption{The progressive search strategy with deep supervision in NAS-ASDet. The search process is decomposed into multiple stages, and operation-level pruning strategy is used to gradually remove operations until generate final architecture.}\label{Fig3}
	\vspace{-0.3cm}
\end{figure}

In this section, we introduce a progressive search strategy with a deep supervision mechanism to explore the search space. As discussed in Section~\ref{subsub2-1}, the original DARTS suffers from extra search costs and rough decoupling. Therefore, we divide the search process into multiple stages, gradually removing operations at each stage, thereby enabling the direct generation of the architecture without additional decoupling. Additionally, inspired by deeply supervised networks, we employ deep supervision to facilitate rapid adaptation to target defects.
This search strategy not only reduces optimization challenges in complex search processes but also contributes to improved detection performance and reduced network design costs.
The entire search diagram is shown in Figure.~\ref{Fig3}.

First, because of the abovementioned continuous relaxation, the search process can be formulated as a nested optimization problem:
\begin{equation} 
	\begin{aligned}
		\label{NAS_define}
		\boldsymbol{\alpha} \gets \min~ Loss_{arc}&(\boldsymbol{\alpha},w_{\alpha}^*)  \\
		s.t.~~ w_{\alpha}^* = \arg\min_{w}~& Loss_{weight}(\alpha,w)
	\end{aligned}
\end{equation}
This indicates that architectural weights $\boldsymbol{\alpha}$ are found to minimize validation loss $Loss_{arc}(\boldsymbol{\alpha},w_{\alpha}^*)$, where $w_{\alpha}^*$ represents the best network weight under the given architecture $\boldsymbol{\alpha}$. The network weight $w$ is updated to $w_{\alpha}^*$ on the architecture training dataset by fixing architectural weight $\boldsymbol{\alpha}$, and then $\boldsymbol{\alpha}$ is updated on the weight training dataset by fixing $w_{\alpha}^*$.

Next, the search process is decomposed into $K$ multiple stages. After each stage of training, operation-level pruning is used to gradually remove operations with small contributions. Specifically, $\boldsymbol{\alpha}$ is used to represent the importance of each operation. At the end of stage training, candidate operations with low importance are removed and $top_k(k=1,\cdots,K)$ important candidate operations are retained. 
\begin{equation}
	\mathcal{O}_k=\mathcal{O}(\boldsymbol{\alpha}^{top\_k})
\end{equation}
where $\mathcal{O}(\boldsymbol{\alpha}^{top\_k})$ represents the shrink search space after $k$-th pruning.

Therefore, the original space $\mathcal{A}(\mathcal{O})$ can be gradually narrowed:
\begin{equation}
	\mathcal{A}(\mathcal{O}_k)\gets \mathcal{A}(\mathcal{O})
\end{equation}
where $\mathcal{O}_k$ denotes the new candidate operation set after pruning according to the sorting of $\boldsymbol{\alpha}$, and $\mathcal{A}(\mathcal{O}_k)$ represents the new search space obtained in the $k$-th stage.

This process is repeated in each stage until each edge remains to be considered in the unique definite operation.

Then, for each intermediate node in the cell, the two strongest operations from different nodes collected from all previous nodes are kept, while other transformations are masked. The strongest operations are defined as follows:
\begin{equation}
	o^{(i,j)}= argmax_{o \in \mathcal{O}} \left( \frac{e^{\alpha_o^{(i,j)}}}{\sum_{o' \in \mathcal{O}}e^{\alpha_{o'}^{(i,j)}}} \right).
\end{equation}

Finally, the determined architecture is searched as:
\begin{equation}
	a \Leftarrow \mathcal{A}(\mathcal{O}_k)\gets \mathcal{A}(\mathcal{O})
\end{equation}

Additionally, deep supervision is applied to reduce the optimization difficulty of the complex search process. This mechanism can be implemented by adding the branch loss to the total training loss:

\begin{equation}
	Loss^{bra}= \sum_{i=1}^{5} Loss^{(i)}, 
\end{equation}
where $Loss^{(i)}$ denotes the branch loss of the $i$-th feature map and $i \in \{1,2,3,4,5\}$. 
The total training loss is defined as the sum of the branch loss and the segmentation loss: $Loss_{total}=Loss^{out}+Loss^{bra} $, where $Loss^{out}$ denotes the segmentation loss between the defect segmentation results and the data labels. In this article, the same loss function as \cite{NDDNet2022} is used as the branch loss, which consists of binary cross-entropy (BCE) and dice similarity coefficient (DICE).

\section{Experiments and Results}\label{sec4}
\vspace{-0.3cm}
In this section, we conduct experiments on industrial datasets to determine:
\textbf{RQ1}: How do the networks designed by NAS-ASDet compare with  manually designed detection networks and other NAS methods?
\textbf{RQ2}: How does the design in NAS-ASdet affect its performance: (a) How advantageous is data-driven multiscale feature extraction compared with classical feature extraction networks? (b) How effective is the adaptive adjustment of the feature importance distribution?
\textbf{RQ3}: How effective is the search strategy used in NAS-ASdet compared to other NAS search strategies?

\vspace{-0.3cm}
\subsection{Experimental Settings}
\subsubsection{Experimental Environment}
We implement our network on PyCharm with the toolbox PyTorch. In the experiments, the method is trained and tested on NVIDIA Tesla A100 (with 40-GB GPU memory), CUDA vision of 11.4, Pytorch vision of 1.9.0, torchvision vision of 0.10.0, and CentOS Linux 8.0.

\vspace{-0.3cm}
\subsubsection{Datasets}
To evaluate the performance of NAS-ASdet, we conduct experiments on four industrial datasets that encompassed various challenges, including lighting conditions, scale variations, limited sample sizes, etc.
\begin{itemize}
\item  \textbf{MCSD-C dataset}: 
This dataset comes from multiple batches of motor commutator cylinder surface defects. Figure.~\ref{Fig4} shows representative surface defect samples that appear on the commutator cylinder surface.
The background environment of MCSD-C is complex and dynamic due to changes in production batches and line parameters. These factors leads to diverse lighting conditions and low-contrast defects, combined with defects exhibiting multi-scale variations, increase the difficulty of defect detection.
We selected 566 defect samples with 256$\times$256.
We used 445 images for training and the remaining samples for testing. Figure.~\ref{Fig5} (1)-(2) rows shows example defect images and corresponding ground truth in MCSD-C.
\begin{figure}[h]%
	\centering
	\includegraphics[width=11cm]{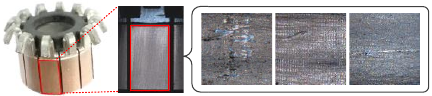}
	\caption{Commutator cylinder surface and its surface defect samples.}\label{Fig4}
\end{figure}

\item  \textbf{RSDDs dataset}\cite{RSDDs2017}:
RSDDs is a rail tread defect dataset collected from general/heavy transportation tracks. The detection challenges include severe surrounding noise interference and limited defect samples with complex contour shapes.
RSDDs contains 128 defect samples with 55$\times$1250.
We augment the original images with a 55-pixel sliding window, and finally obtain 191 training images and 82 testing images. 
Images are resized to 64$\times$64 during training. Figure.~\ref{Fig5} (3)-(4) rows shows example defect images and corresponding ground truth in RSDDs. 
\item  \textbf{KSDD dataset}\cite{Tabernik2019JIM}:
It is captured in a controlled industrial environment with visible surface cracks and contains only 54 defect samples. We augment the defect samples with a 500-pixel sliding window, resulting in 191 training images and 82testing images. KSDD primarily evaluates the capability for low-contrast defects with a limited number of sample images. Images are resized to 512$\times$512 during training. Figure.~\ref{Fig5} (5)-(6) rows shows example defect images and corresponding ground truth in KSDD.
\item  \textbf{DAGM dataset}\cite{wieler2007weakly}:
The artificially created DAGM dataset provides a faithful representation of defects against a textured background. Its main challenge lies in the need to detect multiple types of defects with blurred boundaries under complex backgrounds and textures. In the original DAGM dataset, the defect regions are blanketed roughly by ellipses. In our experiment, four types of defects are selected and redefine the label at the pixel level. Finally we obtain 523 training images and 255 test images, with 512 × 512 original resolution. Figure.~\ref{Fig5} (7)-(8) rows shows example defect images and corresponding ground truth in DAGM.
\end{itemize}

\subsubsection{Baselines}
We compare NAS-ASDet with state-of-the-art handcrafted networks and existing NAS methods.
\begin{itemize}
\item \textbf{Handcrafted network}: First, we select five classical handcraft networks designed for natural scenes as benchmarks for detection performance: FCN \cite{fully2015}, U-Net\cite{ronneberger2015u}, PSPNet\cite{zhao2017pyramid}, DeepLabV3+ \cite{chen2018encoder}, CSNet \cite{cheng2021highly} (salient object detection technique). Secondly, we compare NAS-ASDet with four state-of-the-art manual defect detection networks, including defect segmentation frameworks and detection method based on salient object detection technique: PGA-Net\cite{pga2019},CSEPNet \cite{ding2022cross}, TSERNet \cite{han2022two}, and LSA-Net \cite{li2023lsa}.
\item \textbf{NAS-based method}: 
We compare the NAS-ASDet performance with recent NAS segmentation methods: (a) Auto-DeepLab \cite{liu2019auto}: a macrosearch method for image segmentation, which is a classical baseline for NAS segmentation; (b) NAS-Unet\cite{nasunet2019unet}: which searches for specific cells and achieves good segmentation performance based on cell-based search space; (c) iNAS\cite{gu2021inas}: A hardware-aware saliency detection architecture with high performance based on NAS. (d) DNAS\cite{dnas2022}: which uses the same search space as Auto-DeepLab, with a decoupled search framework to mitigate combinatorial explosion.
\end{itemize}

The above selected comparison methods cover image segmentation techniques, saliency object detection techniques, and lightweight network design, which helps to comprehensively evaluate the performance of our methods.

\subsubsection{Performance Metrics}
We use the same common metrics as \cite{mrd2021} to evaluate model performance, including intersection over union (IoU), F1-Measure (F1), and pixel accuracy (PA). In addition, model Parameters (Params) and floating-point operation per second (FLOPs) are added because industrial sences are more sensitive to resource consumption. We use the total search time (Search\_time) to measure the time consumption in NAS methods.

\subsubsection{Implementation Details}
Given a defect detection dataset, the training process is divided into search and retraining steps.
\begin{itemize}
\item \textbf{Search Stage}: This stage aims to generate deterministic detection architectures from a cell-based search space. The training set is divided 6:4 into an architecture training set and a weight training set, respectively. We define that each cell contains 4 intermediate nodes and 2 input nodes, and the network consists of 4 normal cells and 4 reduction cells. According to the search strategy described in Section~\ref{sub3-3}, we establish 4 search stages, and the original candidate operations (Table~\ref{Tab1}) are gradually pruned in each stage according to $[top_1=7, top_2=4 , top_3=2, top_4=1]$. The maximum epoch and batch size of each stage are set to 70 and 4, respectively. To avoid poor search performance caused by instability at the beginning of a search stage, we only use the weight training set to update network weights $w$ during the first 20 epochs of each stage. In the remaining epochs, architecture weights $\boldsymbol{\alpha}$ and network weights $w$ are updated by alternately using the architecture training set and weight training set. The Adam optimizer with a learning rate of 0.002, weight decay of 0.001 and momentum of 0.9 is used to update $\boldsymbol{\alpha}$, and the SGD optimizer with a learning rate of 0.005, weight decay of 0.0001 and momentum of 0.9 is used to update $w$.
\item \textbf{Retraining Stage}: After the search stage, a deterministic architecture is selected to replace the original supernet. The network weights $w$ are retrained for 500 epochs to mitigate training bias and ensure full network convergence. Multiscale features are fully considered according to the importance distribution. The other hyperparameters are set as in the search stage.
\end{itemize}

\subsection{Performance Comparison (RQ1)}
\begin{figure*}[t]
	\centering
	\includegraphics[width=14cm]{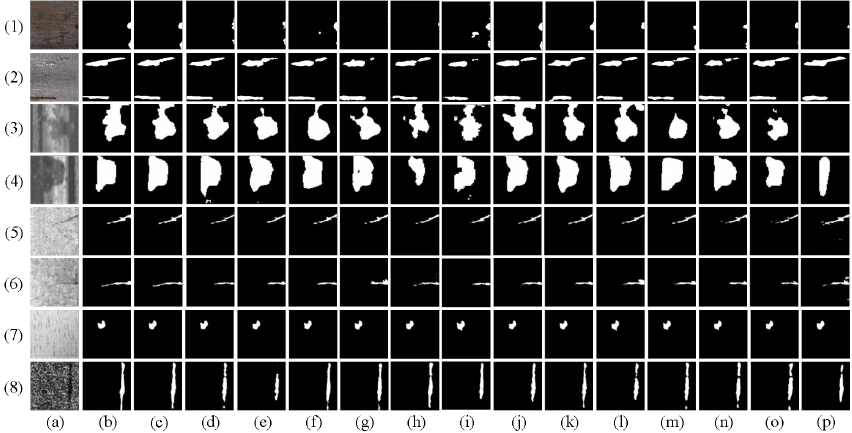}
	\caption{ Comparison of detective results on four different surface defect datasets. (1)-(2): MCSD-C dataset. (3)-(4): RSDDs dataset. (5)-(6): KSDD dataset. (7)-(8): DAGM dataset. (a) Original images. (b) Ground truth. (c) NAS-ASDet. (d) FCN. (e) U-Net. (f) PSPNet. (g) DeepLabV3+. (h) CSNet. (i) PGA-Net. (j) CSEPNet. (k) TSERNet. (l) LSANet. (m) Auto-DeepLab. (n) NAS-Unet. (o) iNAS. (p) DNAS.
	}\label{Fig5}
\end{figure*}

In Table~\ref{Tab2} and Figure.~\ref{Fig5}, we present the quantitative analysis and visual defect prediction results of NAS-ASDet, respectively. Overall, our proposed method outperforms existing approaches in terms of model performance, computational complexity, and visual results, achieving the best detection performance.

\begin{table*}[!h]
	\renewcommand{\arraystretch}{1.5}
	\caption{\centering Performance comparison of different methods on industrial datasets, including handcrafted network s and NAS. \label{tab 2}} 
	\centering 
	\Huge
	\resizebox{\linewidth}{!}{
		\begin{tabular}{*{16}{c}}
			\toprule
			\multirow{2}*{Dataset} & \multirow{2}*{Metrics} & \multicolumn{9}{c}{Manually designing architectures} & \multicolumn{4}{c}{NAS-based architectures} & \textbf{Ours} \\
			\cmidrule(lr){3-11}\cmidrule(lr){12-15}\cmidrule(lr){16-16}\morecmidrules\cmidrule(lr){16-16}
			& & \makecell[c]{FCN\\\cite{fully2015}} &  \makecell[c]{U-Net\\\cite{ronneberger2015u}} & \makecell[c]{PSPNet\\\cite{zhao2017pyramid}} & \makecell[c]{DeepLabV3+\\\cite{chen2018encoder}} & \makecell[c]{CSNet\\\cite{cheng2021highly}} & \makecell[c]{PGA-Net\\\cite{pga2019}} & \makecell[c]{CSEPNet\\\cite{ding2022cross}} & \makecell[c]{TSERNet\\\cite{han2022two}} & \makecell[c]{LSANet\\\cite{li2023lsa}} & \makecell[c]{Auto-DeepLab\\\cite{liu2019auto}} & \makecell[c]{NAS-Unet\\\cite{nasunet2019unet}} & \makecell[c]{iNAS\\\cite{gu2021inas}} & \makecell[c]{DNAS\\\cite{dnas2022}} & \textbf{NAS-ASDet} \\
			\midrule
			\multirow{6}*{MCSD-C} & IoU(\%) $\uparrow$ & 68.53 & 66.13 & 68.10 & 69.17 & 68.66 & 63.61 & \underline{71.79} & 71.04 & 70.28 & 65.70 & 68.13 & 67.13 & 66.54 & \textbf{72.98} \\
			& F1(\%) $\uparrow$ & 79.19 & 77.43 & 79.38 & 80.28 & 82.49 & 75.70 & \underline{83.95} & 81.19 & 80.79 & 77.67 & 79.58 & 78.60 & 78.17 & \textbf{83.39} \\
			& PA(\%) $\uparrow$ & 82.12 & 80.99 & 81.53 & 82.15 & 82.67 & 78.98 & 82.61 & 83.72 & \underline{83.43} & 80.30 & 81.43 & 81.06 & 80.91 & \textbf{84.60} \\
			& Params(M) $\downarrow$ & 32.94 & 7.85 & 46.71 & 39.76 & \underline{0.16} & 51.41 & 18.78 & 189.6 & 25.42 & 3.02 & 1.00 & 6.30 & 8.12 & \textbf{2.43} \\
			& FLOPs(G) $\downarrow$ & 69.43 & 28.20 & 92.44 & 29.88 & 1.01 & 824.53 & 118.66 & 531.83 & 30.77 & 25.44 & 13.66 & \underline{0.85} & 15.93 & \textbf{2.33} \\
			& Search\_time $\downarrow$ & / & / & / & / & / & / & / & / & / & 6:38'19" & 5:13'14" & 9:33'01" & \underline{4:16'19"} & \textbf{4:49'55"}\\	
			\cmidrule(lr){2-16}
			\multirow{6}*{RSDDs} & IoU(\%) $\uparrow$ & 58.11 & 63.62 & 56.61 & 61.43 & 52.85 & 60.85 & 65.42 & \underline{65.63} & 64.07 & 52.02 & 61.09 & 58.29 & 18.33 & \textbf{66.79}\\
			& F1(\%) $\uparrow$ & 71.01 & 75.03 & 69.60 & 73.76 & 69.79 & 73.14 & 76.97 & \underline{77.31} & 76.15 & 65.16 & 73.71 & 70.89 & 25.41 & \textbf{78.21} \\
			& PA(\%) $\uparrow$ & 73.91 & 79.59 & 74.31 & 76.21 & 70.60 & 76.79 & \underline{80.25} & 79.96 & 78.23 & 71.38 & 77.10 & 75.21 & 51.30 & \textbf{80.78} \\
			& Params(M) $\downarrow$ & 32.94 & 7.85 & 46.71 & 39.76 & 0.77 & 51.41 & 18.78 & 189.64 & 25.42 & 5.48 & \underline{0.58} & 5.32 & 22.83 & \textbf{1.27}\\
			& FLOPs(G) $\downarrow$ & 4.33 & 1.80 & 5.88 & 1.87 & 0.15 & 51.53 & 7.41 & 33.24 & 1.92 & 0.79 & 0.55 & \underline{0.04} & 1.82 & \textbf{0.10}\\
			& Search\_time $\downarrow$ & / & / & / & / & / & / & / & / & / & 3:29'40" & \underline{1:42'13"} & 5:5'31" & 1:51'30" & \textbf{1:40'27"}\\	
			\cmidrule(lr){2-16}
			\multirow{6}*{KSDD} & IoU(\%) $\uparrow$ & 64.08 & 68.98 & 65.52 & 68.87 & 64.28 & 67.13 & \underline{70.39} & 68.01 & 69.64 & 69.35 & 69.95 & 64.83 & 65.97 & \textbf{72.78} \\
			& F1(\%) $\uparrow$ & 77.26 & 81.13 & 78.78 & 80.90 & 79.79 & 79.82 & 81.83 & 78.86 & 81.52& 81.62 & \underline{82.01} & 78.23 & 79.04 & \textbf{83.52} \\
			& PA(\%) $\uparrow$ & 80.91 & 83.22 & 80.99 & 81.91 & 80.42 & 81.85 & \underline{84.19} & 82.97 & 83.45 & 82.65 & 83.51 & 81.10 & 80.65 & \textbf{85.24} \\
			& Params(M) $\downarrow$ & 32.94 & 7.85 & 46.71 & 39.76 & \underline{0.69} & 51.41 & 18.78 & 189.64 & 25.42 & 3.10 & 0.71 & 5.09 & 7.60 & \textbf{1.63} \\
			& FLOPs(G) $\downarrow$ & 277.72 & 112.81 & 369.46 & 119.54 & 8.40 & 3298.13 & 474.68 & 2127.32 & 123.09 & 104.51 & 46.67 & \underline{2.34} & 98.04 & \textbf{6.94} \\
			& Search\_time $\downarrow$ & / & / & / & / & / & / & / & / & / & 6:23'10" & 5:35'50" & 7:25'48" & \underline{4:01'59"} & \textbf{2:35'28"} \\	
			\cmidrule(lr){2-16}
			\multirow{6}*{DAGM} & IoU(\%) $\uparrow$ & 68.44 & 75.30 & 77.85 & 79.36 & \underline{80.09} & 77.81 & 79.79 & 79.85 & 79.66 & 78.27 & 79.97 & 76.07 & 76.13 & \textbf{80.66} \\
			& F1(\%) $\uparrow$ & 79.21 & 86.19 & 87.29 & 88.04 & 88.64 & 87.23 & 88.53 & 88.55 & 88.43 & 87.50 & \underline{89.02} & 85.86 & 85.70 & \textbf{89.08} \\
			& PA(\%) $\uparrow$ & 81.51 & 86.65 & 87.80 & 88.71 & 89.13 & 87.86 & 89.02 & 89.10 & 88.93 & 88.16 & \underline{89.14} & 86.98 & 86.71 & \textbf{89.51} \\
			& Params(M) $\downarrow$ & 32.94 & 7.85 & 46.71 & 39.76 & \underline{0.49} & 51.41 & 18.78 & 189.64 & 25.42 & 4.09 & 0.64 & 5.57 & 14.06 & \textbf{2.26} \\
			& FLOPs(G) $\downarrow$ & 277.72 & 112.81 & 369.46 & 119.54 & 8.99 & 3298.13 & 474.68 & 2127.32 & 123.09 & 108.43 & 52.02 & \underline{2.66} & 23.62 & \textbf{7.62} \\
			& Search\_time $\downarrow$ & / & / & / & / & / & / & / & / & / & 39:11’16” & 21:07'27" & 15:59'03" & \underline{10:31'13"} & \textbf{9:28'34"} \\
			\bottomrule
		\end{tabular}\label{Tab2}
	}
\end{table*}

\subsubsection{Quantitative comparison}
First, we begin the analysis with the performance of the designed network.
\begin{itemize}
	\item \textbf{Comparison with handcrafted networks}: As shown in Table~\ref{Tab2}, no fixed handcrafted network consistently achieves the best performance across the four different datasets (the best performance is highlighted by underline, including CSNet, CSEPNet, TSERNet, and LSANet), which demonstrates the necessity of using NAS technology. In contrast, our proposed NAS-based adaptive design method (NAS-ASDet) comprehensively outperforms the existing handcrafted networks and achieves state-of-the-art performance (IoU), both when compared to natural images and defect detection methods. It should be noted that unlike these manually designed networks that require empirical design, our approach adopts an automatic search-based design methodology, which only requires a few hours to automatically obtain the detection network on different datasets, significantly improving network design efficiency.
\end{itemize}

\begin{itemize}
\item\textbf{Comparison with NAS-based design methods}: Our proposed NAS-ASDet consistently outperforms existing NAS-based methods. Particularly, on those datasets with more scarce samples (RSDD, KSDD, MCSD-C), compared with those NAS frameworks based on macro search (e.g. Auto-Deeplab, DNAS), our method shows more stable performance. We can observe an average IoU improvement of over 4\% compared to the existing best-performing NAS methods, which shows the effectiveness of our refined search space. Moreover, we achieve this competitive result in relatively shorter search times. Except for DNAS, which completed the search faster on the MCSD-C dataset, our proposed method designed networks with optimal detection results in the shortest time. Compared to the difference in search performance of DNAS on MCSD-C, our method only takes an additional 33 minutes but achieved 6.44\% IoU improvement. This demonstrates that NAS-ASDet outperforms existing NAS methods in terms of performance.
\end{itemize}

Second, we compare the computational complexity because lightweighting is crucial for  industrial deployment. Table~\ref{Tab2} presents the indicators of computational complexity, including Params and FLOPs. Across the four datasets, the networks' size designed by NAS-ASDet are only about 1M-2M parameters with lower FLOPs. Compared to manually designed networks specifically for defect detection (PGANet, CSEPNet, TSERNet, and LSA-Net), NAS-ASDet utilizes only approximately 10\% of the parameters and 5\% of the FLOPs but achieving state-of-the-art performance. This achievement highlights the advantage of NAS-ASDet in industrial scenarios. Although NAS-ASDet has a slightly larger size than lightweight networks like CSNet and iNAS, the complexity increase is minimal (a maximum of 2M additional parameters and 5G additional FLOPs). While this increase does not significantly impose significant computational burden in industrial applications, but achieving an average IoU improvement of over 5\%.

Conclusively, taking into account comprehensive performance metrics and computational complexity, NAS-ASDet surpasses other competitive methods and achieves state-of-the-art results.

\subsubsection{Qualitative comparison}
In order to visually compare the differences in defect detection performance among different methods, we present the visual results of our proposed method and 13 comparative methods in Figure.~\ref{Fig5}. When those networks designed for natural images (Figure.~\ref{Fig5} (d)-(h) columns) are applied to defect detection, the defect area only be roughly delineated, resulting in blurred details and missing some areas. Although those method designed specifically for surface defect detection (Figure.~\ref{Fig5} (i)-(l) columns) can delineate the defects contour more comprehensively, they often exhibit false positives or false negatives in regions with low contrast features, leading to incorrect defect judgments. When facing the defects of scarce samples and low contrast, the existing NAS methods (Figure.~\ref{Fig5} (m)-(p) columns) seem cannot well overcome the interference such as lighting and noise, which cannot show strong adaptability.
In comparison, NAS-ASDet has more sensitive adaptability. As shown in Figure.~\ref{Fig5} (1)-(2) rows, NAS-ASDet can capture low contrast defect features of different scales under different lighting conditions, without missing any defects. Even with a limited number of training samples and in the presence of environmental disturbances, our method can accurately distinguish the defect contour and obtain prediction results closer to the ground truth shown in Figure.~\ref{Fig5} (3)-(4) rows. Figure (5)-(6) rows demonstrate that NAS-ASDet can accurately detect low contrast defects, even subtle cracks, with limited training samples. And it also shows strong adaptability by sensitively recognizing various types of defects with blurry boundaries in the presence of complex background textures, as shown in  Figure.~\ref{Fig5} (7)-(8) rows.

Therefore, our method can flexibly adapt to the challenge of defect detection, locate defects more accurately, and obtain better visual effects.

\begin{figure*}[!h]%
	\centering
	\includegraphics[width=14cm]{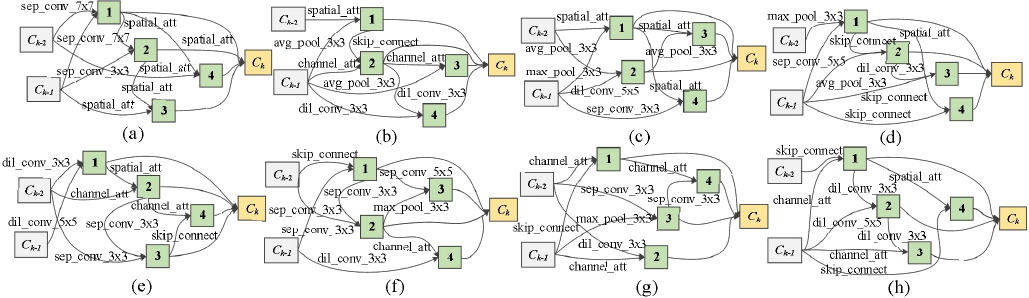}
	\caption{Searched cell architectures on different datasets. (a)(b)  Normal cell and reduction cell searched on MCSD-C. (c)(d) Normal cell and reduction cell searched on RSDDs.  (e)(f) Normal cell and reduction cell searched on KSDD. (g)(h) Normal cell and reduction cell searched on DAGM.}\label{Fig6}

\end{figure*}

To provide a more intuitive representation of the searched network architecture, Figure.~\ref{Fig6} illustrates the searched architectures on different datasets. As expected, the optimal cells obtained vary among these datasets, showcasing the ability of NAS-ASDet to automatically construct cells with diverse architectures driven by data. These architectures are further emphasized by the application of spatial attention and channel attention in different forms to the searched cell, allowing NAS-ASDet to adaptively adjust the feature map's focus and enhance the cell's representational power.

\subsubsection{Failure Case Analysis}
\begin{figure}[h]%
	\centering
	\includegraphics[width=8.5cm]{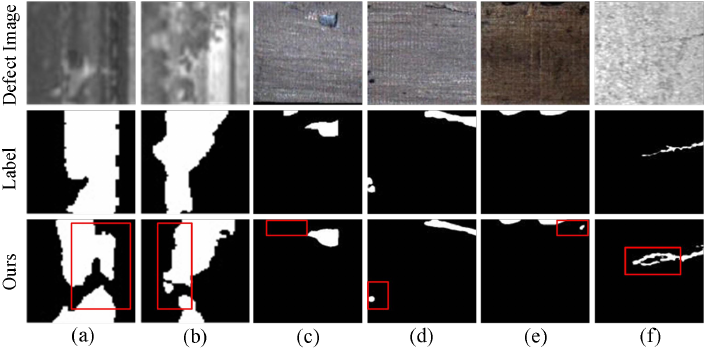}
	\caption{Failure cases of our NAS-ASDet on challenging surface detect images.}\label{Fig7}
\end{figure}

Although NAS-ASDet outperforms other competitive methods, there are still failure cases in some challenging situations. In Figure.~\ref{Fig7}, (a) and (b) columns illustrate typical defect regions with large proportions. However, NAS-ASDet may not consider the entire defect, resulting in partial loss when the defect area changes. For small defects with low contrast at the edge, as shown in the (c) and (d) columns of Figure.~\ref{Fig7}, our method may miss detection. In the (e), (f) column of Figure.~\ref{Fig7}, our method may overly focus on the vicinity of the defect area, leading to false detection in nearby regions. The main reason for this issue is the dataset availability and diversity, and we will address these deficiencies in our future work.

\subsection{Ablation study (RQ2)}
To study how designs in NAS-ASDet affect performance, we design a series of ablation experiments for evaluation.

\subsubsection{Data-driven multiscale feature extraction}
In NAS-ASDet, feature extraction is performed by repeatedly searching cells. We compare this approach with widely used classical feature extraction methods (VGG16 \cite{simonyan2014very}, ResNet50 \cite{he2016deep}, GoogLeNet \cite{szegedy2015going}, MobileNetV3
\cite{howard2019searching} and DenseNet \cite{huang2017densely}). During the comparison, we separately replace the feature extraction stage, while the other components of NAS-ASDet remain intact. As shown in Table~\ref{Tab3}, with the fixed connection mode, the extracted features are fixed and the performance is limited. Our search method can thus automatically determine a more suitable connection, making the network performance is better. Moreover, the searched architecture is more lightweight compared to these classical fixed feature extraction methods.

\begin{table}[!hbtp]
	\caption{\centering Quantitative comparisons of different feature extraction methods on industrial datasets} 
	\centering 
	\renewcommand{\arraystretch}{1.5}
	\vspace{-0.2cm}
	\resizebox{79mm}{24mm}{
		\begin{tabular}{*{5}{c}} 
			\toprule 
			Methods & \multicolumn{2}{c}{RSDDs} & \multicolumn{2}{c}{MCSD-C}\\
			\midrule
			&	IoU(\%) &	params &	IoU(\%) &	params\\
			VGG16 \cite{simonyan2014very}&	66.79 &	15.0M&	55.50&	15.0M\\
			ResNet50 \cite{he2016deep}&	66.78&	25.4M&	47.71&	25.4M\\
			GoogLeNet \cite{szegedy2015going} &	69.14&	7.5M&	49.21&	7.5M\\
			MobileNetV3 \cite{howard2019searching}&	63.61&	2.6M&	42.32&	2.6M\\
			DenseNet \cite{huang2017densely}&	69.02&	9.0M&	49.86&	9.0M\\
			\midrule
			\textbf{Ours} & \textbf{72.98} & \textbf{2.4M} & \textbf{66.79} & \textbf{1.3M}\\
			\bottomrule  
		\end{tabular}\label{Tab3}}
		\vspace{-0.5cm}
\end{table}

\subsubsection{Adaptive adjustment of feature distribution}
\begin{figure}[h]%
	\vspace{-0.1cm}
	\centering
	\includegraphics[width=10cm]{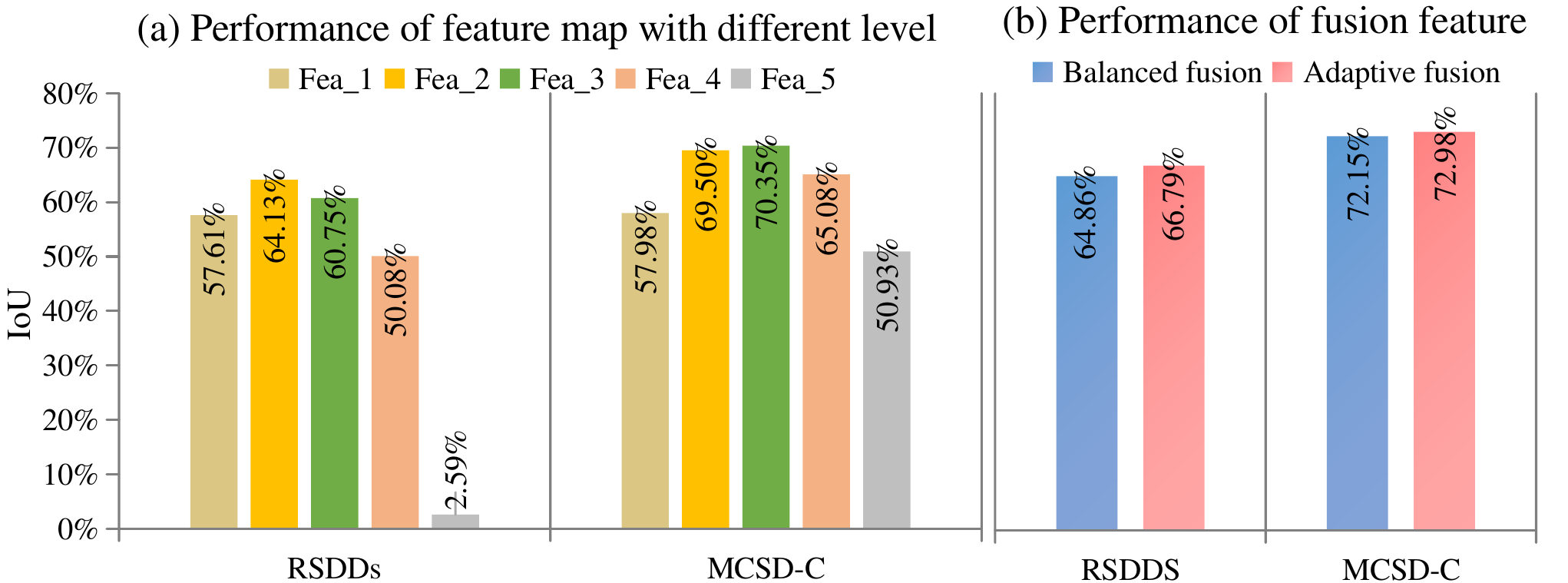}
	\caption{Detailed performance of different multiscale features.}\label{Fig8}
\end{figure}

To consider the impact of fusing multiscale features, we evaluate different levels of feature maps ($Fea_1, Fea_2, Fea_3, Fea_4, Fea_5$) extracted by automatically searched cells. 
We also compare the performance of balanced fusion and adaptive fusion.
As shown in Figure.~\ref{Fig8}:
(1) The detection ability of fusing multiscale features is better than that of single-level features (Figure.~\ref{Fig8}(b) VS Figure.~\ref{Fig8}(a)), which proves that multiscale feature fusion can improve the detection performance;
(2) Different single-level features show different performances (Figure.~\ref{Fig8}(a)). For example, $Fea_2$ achieves the best performance on RSDDs, while $Fea_3$ performs best on MCSD-C. This proves that the contributions of each level of features are different for different tasks, and it is necessary to integrate different scale features according to their importance so that the total performance can be enhanced. (3) The experiments show that adaptive fusion not only outperforms single-level features but also outperforms balanced fusion (Figure.~\ref{Fig8}(b)), proving that the adaptive adjustment of feature importance distribution is effective.

\subsection{Search Efficiency (RQ3)}
We evaluate the search efficiency from the perspectives of search time and model performance. We compare the search strategy in NAS-ASDet with the original DARTS and related improved DARTS, including DARTS \cite{darts2018}, PC-DARTS \cite{pcdarts2019}, P-DARTS \cite{pdarts2019}, and Att-DARTS \cite{nakai2020att}. The same search space as considered with NAS-ASDet is used (to ensure fair comparison). The maximum epoch remains the same as NAS-ASDet, which is set to 280 (70 epochs$\times$4). For the nonstaged search strategy, in the first 80 epochs (20 epochs$\times$4), only the network weight $w$ is updated to avoid initial instability.

\begin{figure}[h]%
	\centering
	\includegraphics[width=9cm]{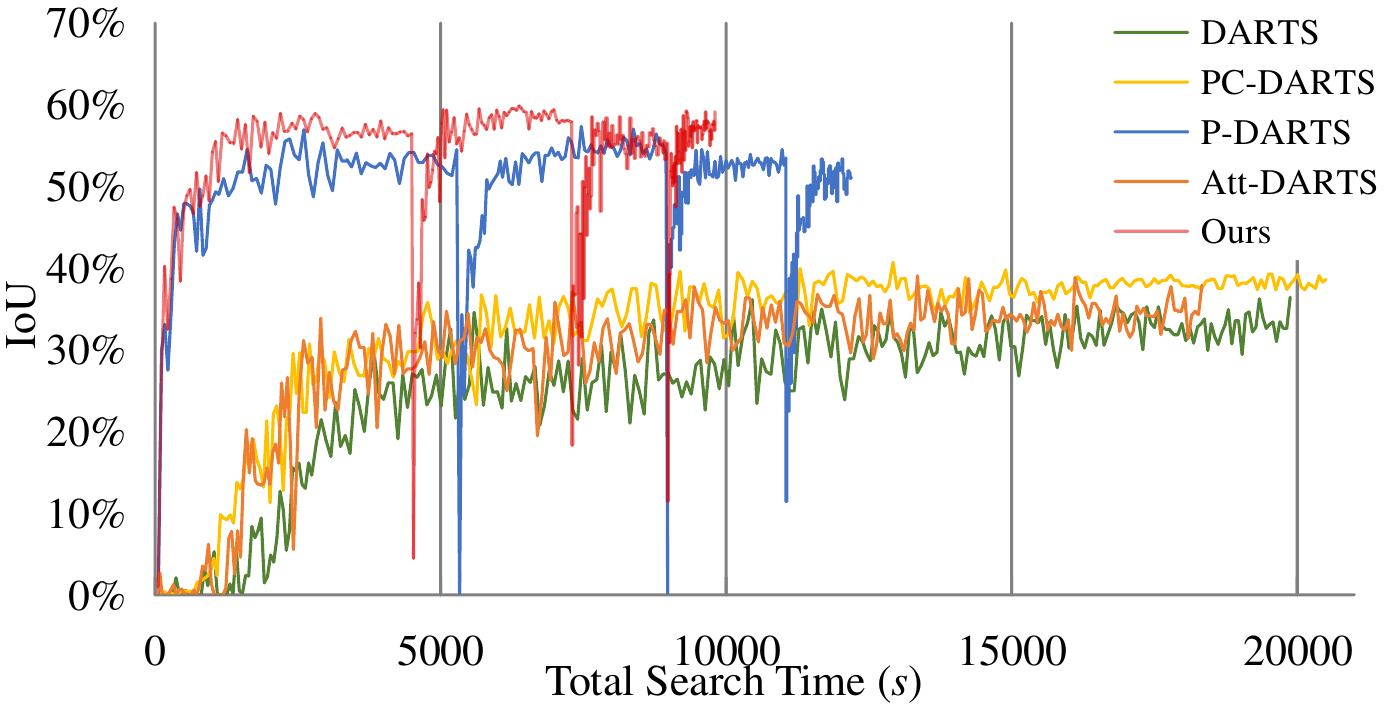}
	\caption{Search time (in seconds) and performance comparison of different search strategies on RSDDs.}\label{Fig9}
\end{figure}

The search efficiency comparison results are shown in Figure.~\ref{Fig9}. The search time of the proposed method is less than that of the others, while the architecture with better performance is generated. Under the condition of limited time resources, our method can generate favorably performing architectures in a shorter time. In summary, the proposed method can explore the NAS-ASDet search space better and faster.

\section{Conclusion}\label{sec5}
In this article, we propose an NAS-based method for adaptive architecture generation in surface defect detection, NAS-ASDet. First, we design a refined and industry-appropriate lightweight search space based on prior manual architecture knowledge. Second, we introduce a cell architecture with data-driven capability and searchable attention operations. Additionally, we desgin a multi-scale feature fusion that can adaptively adjust feature distribution. Furthermore, a progressive search strategy with deep supervision is  designed to effectively explore the search space. Experimental results demonstrate that NAS-ASDet outperforms both manually designed architectures and NAS ones.

In future work, we plan to add more degrees of freedom for architecture search (e.g., scalable architecture) to further reduce the expressive limitation caused by search space. At the same time, we will try to improve the inference speed by using hardware-aware NAS. Additionally, implementing an effective data augmentation strategy combined with our method is another future direction.

\bibliographystyle{elsarticle-num} 
\bibliography{reference}

\end{document}